\documentclass[10pt]{iopart}
\usepackage{graphicx,lipsum,multicol}

\usepackage{iopams}  
\usepackage{orcidlink}
\begin{document}
\title{Bionic Collapsible Wings in Aquatic-aerial Robot}
\newcommand{\orcidauthorA}{0000-0003-0034-9037} 
\author{ 
 Xiao Xiong~\orcidlink{0000-0002-8308-6270}\& Xinyu Zhang~\orcidlink{0000-0003-0034-9037} \&Huanhao Huang~\orcidlink{0000-0001-8190-0390}\&Kangyao Huang~\orcidlink{0000-0002-5708-2620}}

\address{The School of Vehicle and Mobility, Tsinghua University, Beijing,
P.R.China.}
\ead{xiongxiao917@gmail.com}
\vspace{10pt}
\begin{indented}
\item[]April 2023
\end{indented}

\begin{abstract}
 The concept of aerial-aquatic robots has emerged as an innovative solution that can operate both in the air and underwater. Previous research on the design of such robots has been mainly focused on mature technologies such as fixed-wing and multi-rotor aircraft. Flying fish, a unique aerial-aquatic animal that can both swim in water and glide over the sea surface, has not been fully explored as a bionic robot model, especially regarding its motion patterns with the collapsible pectoral fins. To verify the contribution of the collapsible wings to the flying fish motion pattern, we have designed a novel bio-robot with collapsible wings inspired by the flying fish. The bionic prototype has been successfully designed and fabricated, incorporating collapsible wings with soft hydraulic actuators, an innovative application of soft actuators to a micro aquatic-aerial robot. We have analyzed and built a precise model of dynamics for control, and tested both the soft hydraulic actuators and detailed aerodynamic coefficients. To further verify the feasibility of collapsible wings, we conducted simulations in different situations such as discharge angles, the area of collapsible wings, and the advantages of using ground effect. The results confirm the control of the collapsible wings and demonstrate the unique multi-modal motion pattern between water and air. Overall, our research represents the study of the collapsible wings in aquatic-aerial robots and significant contributes to the development of aquatic-aerial robots. The using of the collapsible wings must a contribution to the future aquatic-aerial robot.
\end{abstract}

\submitto{\BB}
\maketitle
\ioptwocol

\section{Introduction}

In recent years, the aquatic-aerial robot has gained widespread attention as an emerging concept that offers a convenient and effective solution for water-air cross-domain tasks. With the ability to move both in air and water. These robots can operate in a wide range of environments, including oceans, pools, and complex urban settings. As a result, they hold great potential for environmental and scientific applications, such as water health monitoring\cite{rf1} and animal behavior observation\cite{rf2}. 

Mature technologies like fixed-wing robots, multi-rotor robots and multi-modal robots\cite{rf3} are used as the flying part of the most aquatic-aerial robots. Fixed-wing robots use fixed wings to generate lift and can fly at speeds above 20-30 $m/s$, allowing them to be deployed over longer distances. Examples of fixed-wing aquatic-aerial robots include those developed by North Carolina State University\cite{rf4} and Dipper\cite{rf5}. Multi-rotor robots, on the other hand, use a structure similar to that of multi-rotor aircraft, with four propellers controlling the robot's motion. The main advantages of multi-rotor robots are their unique hovering and vertical take-off and landing capabilities. Examples of multi-rotor aquatic-aerial robots include the one developed by Beihang University\cite{rf6}. Additionally, some aquatic-aerial robots combine both fixed-wing and multi-rotor structures, such as Nezha III\cite{rf7}, which has the ability to perform horizontal flight, vertical flight, and underwater gliding. Each type of flying method of aquatic-aerial robot has its own unique feature. 

In addition, bionic technology provides other methods to design the aquatic-aerial robot, especially inspired by creatures with the capacity of moving both in the air and water like sea birds, flying squid and flying fish. For instance, based on the biological gannet, a kind of sea birds, the robot developed by Beihang University focuses on wing load investigations during plunge-diving\cite{rf8}. Similarly, the AqualMAV is also inspired by plunge-diving birds and has been tested for performance\cite{rf12}. Because of the plunge-diving method to move in the water for sea birds, a common concern of this kind robot is the huge forces that occur when they impact the water, which may lead to structural damage. In contrast, marine creature-inspired robots tend to avoid high-speed impacts with the water and focus more on the motion in the air. For instance, the robotic jet thruster of AqualMAV was inspired by the flying squid, which uses pressurized water to achieve high speeds during conversion\cite{rf13}. Another squid-like aquatic-aerial vehicle also uses a similar thrust mechanism but emphasizes bionic mechanisms such as soft morphing fins and arms\cite{rf14}. Flying fish have also inspired the design of aquatic-aerial robots. For example, the Flying Fish ocean surveillance platform developed by the University of Michigan avoids hydrodynamic drag from ocean waves and currents while flying, but only has locomotion on the water surface without underwater movement\cite{rf15}. MIT has detailed design considerations for a robotic flying fish prototype\cite{rf16}. 

The locomotion capacity in the air of aquatic-aerial robots base on the mature technologies or inspired by sea birds are active with a power take-off system. Multi-rotor, propeller and flapping wings all consume the energy in the air, while aquatic-aerial robots inspired by the marine creatures do not. These marine based robots usually take off out of water with a fast initial velocity and finish the air motion consuming the initial kinetic energy. Flying squid and flying fish are two typical animals move in the air without power output. However, there are still some difference in air motion between flying squid and fling fish. Flying squid can only drop back to the water by gravity with a parabola trajectory without controlling. Flying fish is a kind talented animal, which can control the gesture and motion state with their pectoral fins and body center in the air. The flexible locomotion with no power need contributes to the unique motion pattern of the flying fish. We curious about this unique motion pattern of flying fish and concentrate on the its collapsible pectoral fins.  

%Flying fish is a typical creature that can move both in the air and water. It is believed that flying fish evolved their flying capacity to evade capture by predators in the water. Their long wing-like pectoral fins and narrow body are their physical characteristics. The pectoral fins, which range from $60-80\%$ of their standard length, generate enough lift force to support the gliding process in the air. The location of the center of gravity, which is located near the base of the pectoral fins, is also a significant factor in the gliding stability\cite{rf9}. 

\begin{figure*}[h]
\centering
\includegraphics[width=17cm]{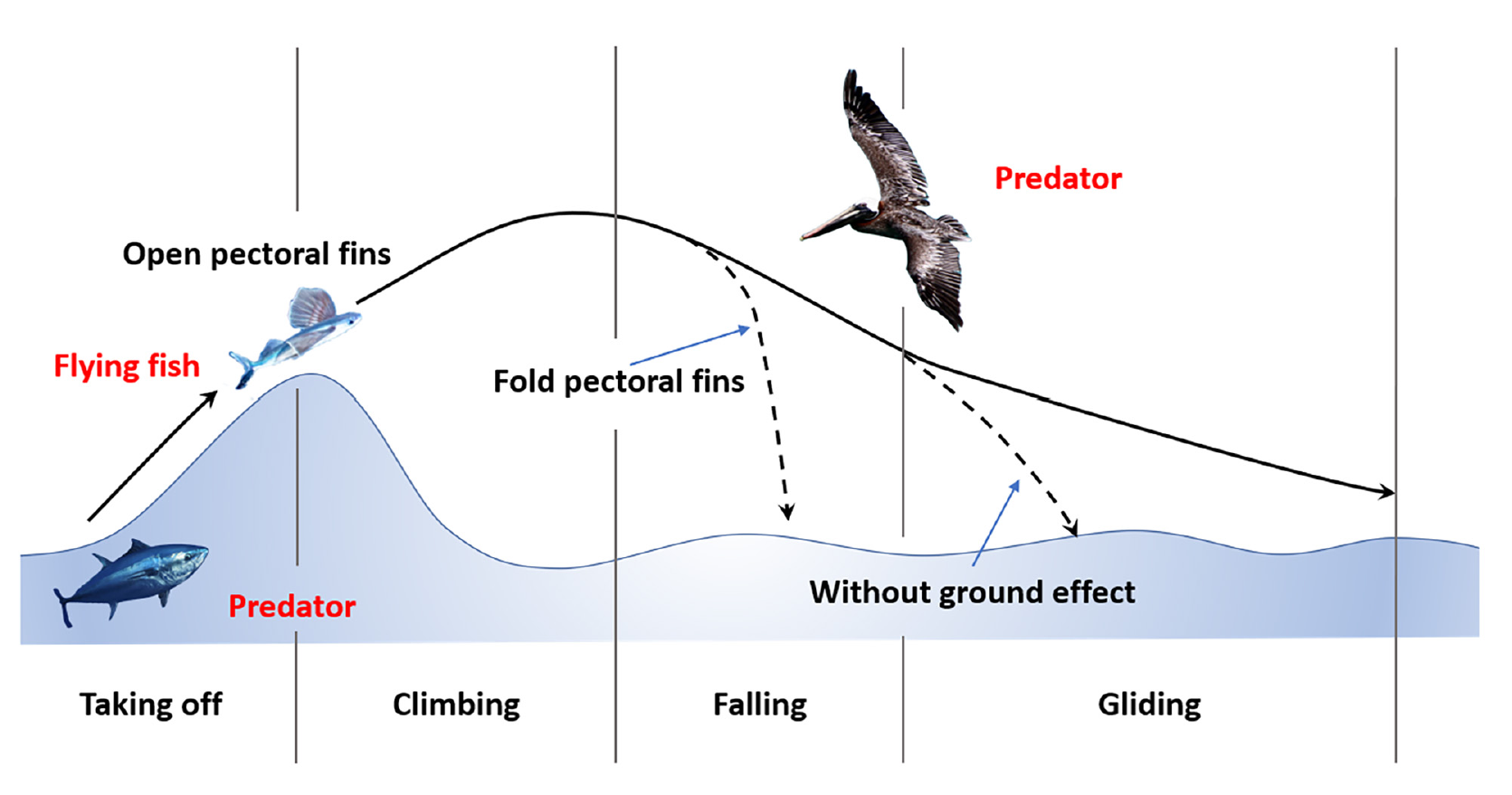}
\caption{Overview of flying fish motion pattern.}. 
\label{flyingfish}
\vspace{-5 mm} 
\end{figure*}

The process of flying fish moving in the air consists of four stages, including taking off, climbing, falling, and gliding stages. During the taking off stage, flying fishes first accelerate underwater. Once they reach a certain speed, they change their forward direction upwards and fly out of the water at maximum speed. Also, they accelerate on the incline water surface and take off at the top of the wave in this stage. After taking off, flying fishes open their pectoral fins to create lift, resist gravity, and glide in the air. Besides, When sliding down the water surface, they only need to swing their tail fins into the water to accelerate again, regain high speed, and take off again. In the gliding stage, fully utilizing of the ground effect contributes to further distance movement\cite{rf10}. It's said that flying fishes can continue this process in the air for a total distance of 400m. In addition, flying fishes have the capacity to change the motion gesture and statement by controlling the pectoral fins. For example, they fold their fins and plunge into the water to escape when they are chased by sea birds and adjust their attack angle of the fins to generate a higher lift force when drop in the falling stage.

Fig.\ref{flyingfish} shows the motion pattern of the flying fish in the air with stages. The solid line represents the trajectory of the flying fish during the entire flying process, while the two dotted lines represent the trajectory when the flying fish folds its wings in the air and the hypothetical situation without the ground effect. Overall, the unique motion pattern of the flying fish serves as an excellent model for designing aquatic-aerial robots that can move efficiently in both air and water.

It's obviously that the collapsible pectoral fins contribute to all the motion in the air. Whether in lift generating or motion controlling in the air, collapsible fin is the main actuator even the only one. The advantage of the collapsible wings like pectoral fins can contribute to the design of the new aquatic-aerial robots. However, there are seldom research focus on the mechanism of the pectoral fins of the flying fish. Thus, further study about the elements effecting the performance of the collapsible need to be done.

%The locomotion of animals has been perfected through long-term evolution. Diving birds and flying fish are two main multi-functional aquatic-aerial creatures that can fly, dive, and swim\cite{rf2}. Diving birds fly in the air and dive in the water for food, and the range of their dives can be between 2 and 5.9$m$. The folded wings and streamlined structures of diving birds enhance their ability to move efficiently through both air and water. Gannets, for example, are a type of plunge-diving bird that can dive from 30$m$ at speeds exceeding 25$m/s$. However, the high speed of their dives creates a huge impact with the water, which is equivalent to approximately 7g (where "g" is the gravitational acceleration)\cite{rf8}.

In this study, The novelty of this study include: (1) a new design of the aerial aquatic robot with collapsible wings inspired by flying fish, (2) the use and precise control of the soft hydraulic actuator replace the rigid actuator of the collapsible wings, (3) a precise dynamic modal of the robot with collapsible wings, and (4) a series of simulation experiments focus on the collapsible wings, studying the effect of discharge angle, area and ground effect to the wing performance.  

This paper is organized as follows. Section II introduces the design and fabrication of the prototype of robot and collapsible wings. Section III analyzes the dynamics of the robot in multi-modal motions. Section IV presents the experimental results and discussions regarding the locomotion of the robot, with a focus on the collapsible wings. Finally, Section V concludes the paper.

\section{Design and Fabrication of Robot Prototype }

\subsection{General Overview }

The unique movement pattern of flying fish is determined by its unique physiological structure especially the collapsible wings. We design an aquatic-aerial robot prototype to study the multi-modal motion with collapsible wings. 

The robot was designed with the appearance and physiological mechanism of the flying fish in mind, with a long and narrow main body and large collapsible pectoral fins that occupy 70\% of the projected area of the body when expanded. To simplify the complexity of the tail fin, a high-powered motor was used as the source of power to accelerate. In addition, two pelvic fins were designed to balance the posture and movement of the robot underwater.

\begin{figure}[t]
\centering
\includegraphics[width=8.6cm]{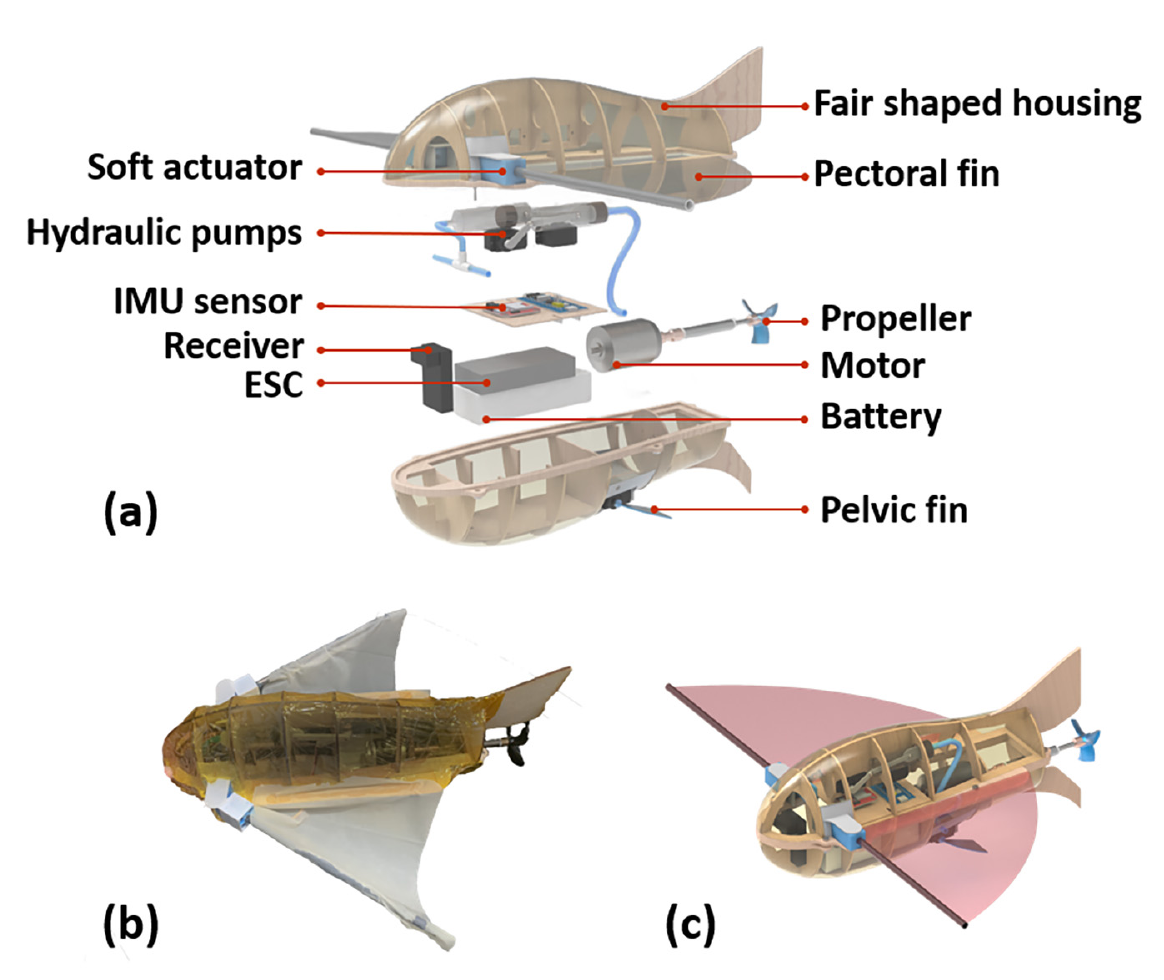}
\caption{ \textbf{(a)} is the exploded view of robot prototype; \textbf{(b)} is the picture of the prototype; \textbf{(c)} is the modal of the prototype. }
\label{explod}
\vspace{-5 mm} 
\end{figure}

The robot is comprised of a streamlined external structure, collapsible wings with soft hydraulic actuator, high power density drive system, and precise control system. The appearance and biological parameters of the flying fish were both referenced to restore the structure, resulting in a robot prototype that closely resembles its natural counterpart. Fig. \ref{explod} shows an exploded view and pictures of the robot prototype, while Table \ref{chart1} provides details on the physical characteristics of the robot.

\begin{table}[]
    \centering
    \caption{Summary of key design parameters}
    \begin{tabular}{c c}
    \br
    Parameter  & Value \\
    \mr
    Length  & 336.1mm\\
    Wing span&455mm\\
    Width(wing folded)&135.4mm\\
    Motor power&300w\\
    Weight &353g\\
    \br
    \end{tabular}
    \label{chart1}
\end{table}

\subsection{Bionic desgin of the main body}

Aquatic-aerial robots need the characteristics of both underwater and aerial robots, such as Waterproof and weight limit. Besides, the streamlined shape with excellent hydromechanical properties is significant for both types of robots. The shell frame of the robot is made of basswood, linked with mortise and tenon joints, reducing the weight without using heavy metallic connectors like bolts. The outer surface is covered with a heat-shrink film, which shapes the streamlined surface perfectly by heating and provides great water-proofing ability. A yellow polyimide film is pasted on the surface to enhance the hardness for scratch prevention in the uncertain underwater environment. The vertical flat and elongate body shape similar to fish, along with two upper and lower tail fins, prevents rotation in the water.

The control system is installed inside the housing, and it is comprised of a motor with a propeller that generates thrust, a hydraulic system driven by servos with a connecting rod structure to control the collapsible wings and the density of the robot for floating or submerging. A receiver receives signals from the remote control, and IMU sensors, pressure sensors, and other sensors connecting to the STM32 chip make up the information collection system. A lithium battery with 11.1V and 1000mAh provides energy for the whole system, which is distributed to all actuators by an electronic speed controller. 

%(\begin{figure}[t]
%\centering
%\includegraphics[width=8.6cm]{controlframe.eps}
%\caption{Control system frame }. 
%\label{Controlsystem}
%\vspace{-5 mm} 
%\end{figure}

\subsection{Collapsible wings with soft actuator}
\begin{figure}[h]
\centering
\includegraphics[width=8.7cm]{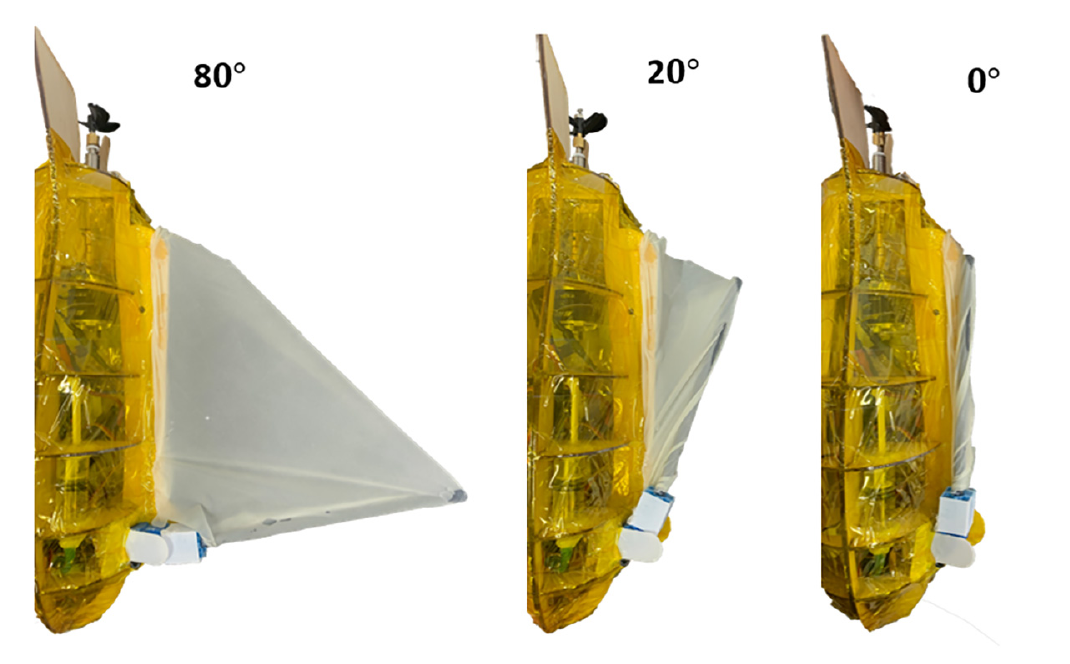}
\caption{Different angle of collapsible wings. }. 
\label{areoofwings}
\vspace{-5 mm} 
\end{figure}

The ability to extend and retract their fins is a critical feature that enables them to adapt to the different environments of air and water. And the collapsible wings in the aquatic-aerial robot are a critical feature that imitates the flying fish mechanism. The use of a soft actuator to control the pectoral fins' extending and retracting is a significant improvement over the rigid rotating structure of most collapsible wings. The soft actuator achieve the precise control in both air and water and have ability to adapt to external forces while maintaining control. The flexible membrane used in the wings of the robot has excellent tensile properties and hydrophobic ability, which reduces weight gain and improves performance in different environments. Fig.\ref{areoofwings} shows the collapsible wings in different open angles.

The use of soft actuators in the collapsible wings of the robot is an innovative approach to mimic the mechanism of flying fish. Unlike rigid rotating structures that are heavy and complex, the soft actuator is light and simple, making it more suitable for use in compact spaces. The injection of liquid into the actuator controls the bending angle, and the internal pressure can be directly controlled by the amount of material volume injected. The flexible capacity of the actuator also allows it to adaptively deform in the face of external impacts, preventing damage due to huge forces. This design not only mimics the natural mechanism of flying fish but also provides a practical and efficient solution when wings of aerial robot get an impact in the movement.

\begin{figure}[h]
\centering
\includegraphics[width=8.7cm]{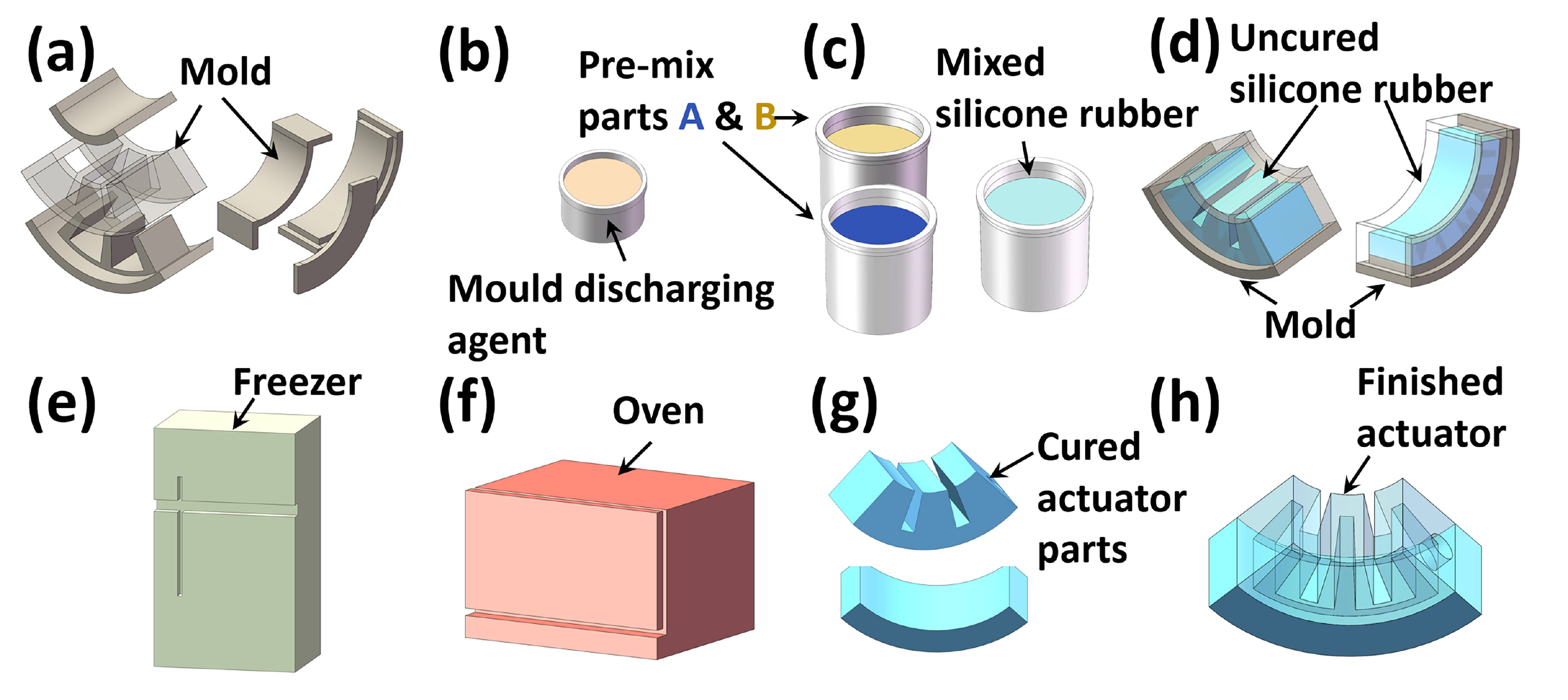}
\caption{Fabrication of the soft actuator. \textbf{(a)} two actuator mold parts \textbf{(b)} applying mould discharging agent to the inner surface \textbf{(c)} mixing and stirring the pre-mix silicone rubber evenly \textbf{(d)} pouring uncured silicone rubber into the molds \textbf{(e)} refrigerating to remove air bubbles \textbf{(f)} curing the silicone rubber at temperature(140$^{\circ}$F/60$^{\circ}$C) in the oven \textbf{(g)} getting upper and lower parts of actuator \textbf{(h)} gluing the two parts through secondary pouring}. 
\label{flowchart}
\vspace{-5 mm} 
\end{figure}

Soft actuator is designed by the modal of soft hydraulic actuator. Not similar to the normal straight hydraulic actuator at beginning, a 90 degree bending actuator was designed to suit the situation that the wings is retracted without driving. With the injection of the liquid, the actuator formation is straightened to open the wing. The bending angle is decided by the inner pressure.  Fig.\ref{flowchart} is the process of actuator fabrication and the silicone rubber is Mold Star 30, Smooth-on, Inc, Easton Pa, USA.

To increase the size of the wings without affecting the robot's propulsion system, a 170mm carbon fiber tube was used as the arm of the wing to support the flexible membrane at the end of the actuator. The membrane of the fin was made of a 0.07mm hydrophobic elastic silicone film, which had excellent tensile properties. When deployed, the membrane formed a large airfoil, but when retracted, it was almost negligible. The hydrophobic properties of the membrane's surface also reduced weight gain caused by water attachment after the robot left the water.

\section{Dynamics Analysis}
In order to comprehend the effect of collapsible wings and the motion of the aquatic-aerial robot inspired by the flying fish, dynamics of the robots motion must be calculated and analysed. In this part, coordinate system of robot has been built, and forces and torques in three different stages has been calculated.  
\subsection{Coordinate System and kinematic}
\begin{figure}[h]
\centering
\includegraphics[width=8.7cm]{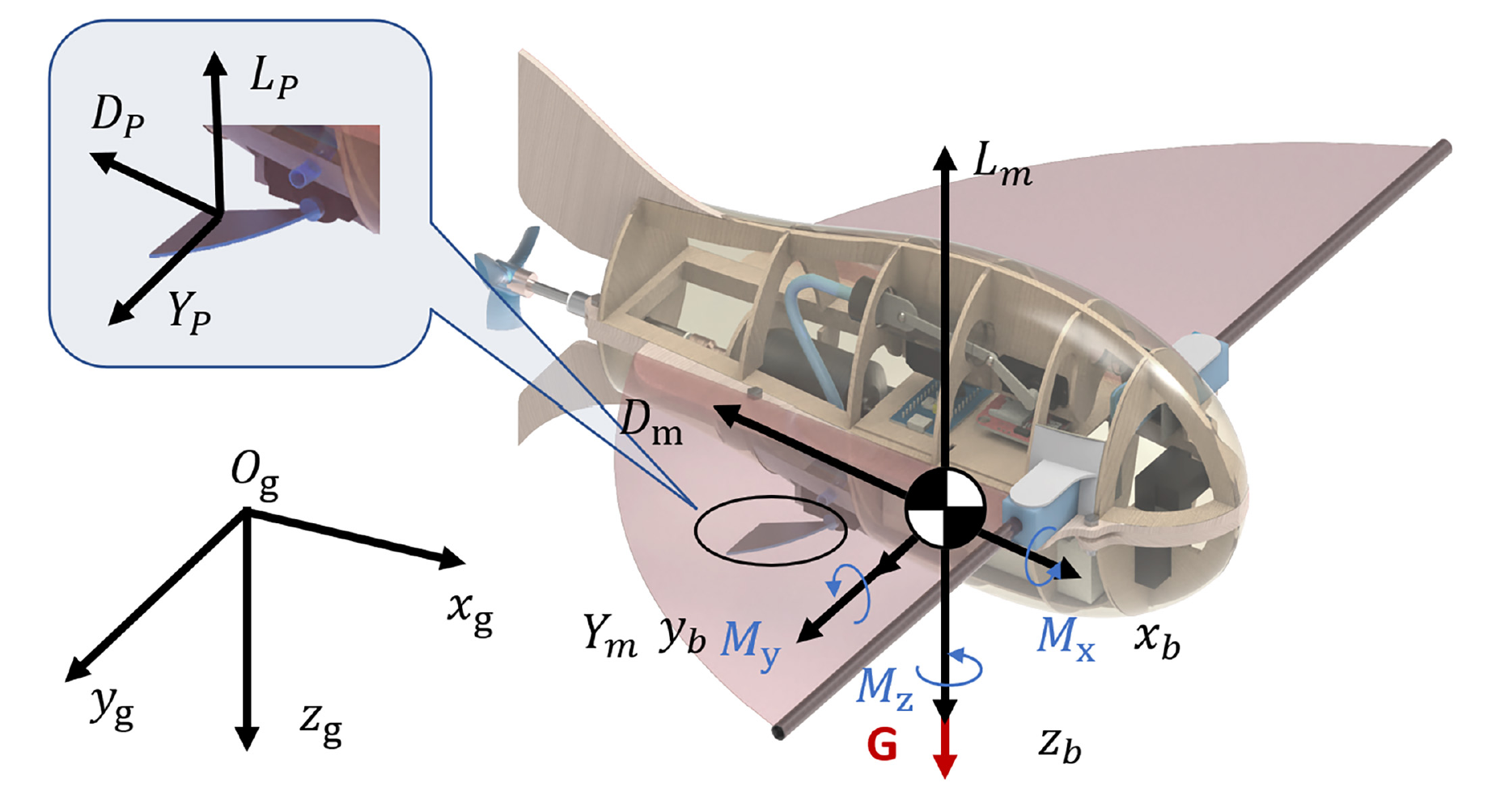}
\caption{Coordinate system.}
\label{Frame}
\vspace{-5 mm} 
\end{figure}
The robotic schematic with three coordinate frames were established and  showed in figure \ref{Frame}, including the body frame $O_{body}$ , ground frame $O_{ground}$  and velocity frame $O_{velocity}$. The body frame $O_{body}$  established on the centroid of the robot. the $x_{b}$-axis points to the front of the robot and the $z_{b}$-axis point vertically down of the robot body. The $y_{b}$-axis is determined by the right-hand rule. These three frames rotate to each other as Eq.\ref{con:eq1}:

\begin{equation}
    \eqalign{O_{velocity}=R_{vb} O_{body}\\
    O_{body}=R_{bg} O_{ground}}\label{con:eq1}
\end{equation}

\begin{equation}
\eqalign{R_{vb} =
\left[
\begin{array}{ccc}
    C_{\alpha} C_{\beta} & S_{\beta} & S_{\alpha} C_{\beta}  \\
    C_{\alpha} C_{\beta} &  C_{\beta} & -S_{\alpha}S_{\beta}\\
    -S_{\alpha} & 0 & C_{\alpha}
\end{array}
\right]\\
R_{bg}=
\left[ \begin{array}{ccc}
     C_{\theta} C_{\psi} & S_{\theta} S_{\psi} & -S_{\theta}\\
    S_{\theta} C_{\psi} S_{\emptyset}-S_{\psi} C_{\emptyset} & S_{\theta} S_{\psi} S_{\emptyset}+C_{\psi} C_{\emptyset} & C_{\theta} S_{\emptyset} \\
S_{\theta} C_{\psi} C_{\emptyset}+S_{\psi} S_{\emptyset} & S_{\theta} S_{\psi} C_{\emptyset}-C_{\psi} S_{\emptyset} & C_{\theta} C_{\emptyset} 
\end{array} 
\right]}
 \end{equation}

\noindent where,$\alpha$ is the angle-of-attack and $\beta$ is the side-slip-angle. The $\theta$, $\psi$ and $\emptyset$ are the angles of pitch, roll and yaw, respectively. $C$ and $S$ represent trigonometric functions sine and cosine.

 Angular velocity vector and translational velocity vector are recorded as $\Omega=[p\  q\  r]'$ and  $V=[u\  v\  w]'$. The motion equations of robot in the body frame are shown as:
\begin{equation}
    \dot{\Omega }= I^{-1}\left ( M-\Omega \times I\Omega  \right )
\end{equation}
\begin{equation}
    \dot{V}= \frac{F}{m}-\Omega \times V,
\end{equation}
where $m$ is the mass of robot and $I$ is the inertia matrix. Two vectors F and M represent force and torque.Then introduce the vectors $\Phi=[\phi\  \theta\  \psi]'$ and $X=[x\  y\  z]'$, which respectively represent the Euler angle and the center of gravity position of the flying fish. Both are representations of the ground coordinate system. The kinematic equations and navigation equations are:
\begin{equation}
    \dot{\Phi} ={S_{bg}}^{-1}\Omega ,\dot{X} ={S_{ag}}^{-1}V 
\end{equation}

\subsection{Forces and Torques in the movement}
\subsubsection{Gliding}
When gliding in the air, the center of gravity moment M is the result of considering the engine thrust, ground effect, and the aerodynamic force of the pectoral and pelvic fins of the flying fish. It can be expressed as:
\begin{equation}
  M=M_{m}+M_{p}=\frac{1}{2}\rho V_{m}^{2}S_{1}
\left[\begin{array}{ccc}
  bC_{x}
\\ cC_{y}
\\bC_{z}   
\end{array}
\right]
+\frac{1}{2}\rho V_{p}^{2}S_{1}
\left[\begin{array}{ccc}
 bC_{x}
\\ cC_{y}
\\bC_{z}   
\end{array}
\right]
\end{equation}
$M_{m}$, $M_{p}$ are the moments generated by the aerodynamic force of fins. $C_{x}$, $C_{y}$ and $C_{z}$ are the roll, pitch, and yaw moment coefficients of the flying fish, respectively. $\rho$ is air density, $V_{m}$, $V_{p}$ are center of gravity velocity and pelvic fin velocity respectively. The speed of the pelvic fin is:
%$M_{m}$ is the moment generated by the aerodynamic force of pectoral fin. $C_{x}$,$C_{y}$ and $C_{z}$ are the roll, pitch, and yaw moment coefficients of the robot, respectively. $\rho$ is air density, $V_{m}$ is the center of gravity velocity 
\begin{equation}
    V_{p}=V_{m}+R ( \frac{\pi }{2} -\theta _{p})(\dot{\theta _{p}}\times b_{p}),
\end{equation}
where $R$ is expressed as a matrix rotating around the $Y_{b}$ axis, $\theta_{p} $is the pitch angle of the pelvic fin, and $b_{p}$ is the average width of the pelvic fin.
\begin{figure}[h]
\centering
\includegraphics[width=8.6cm]{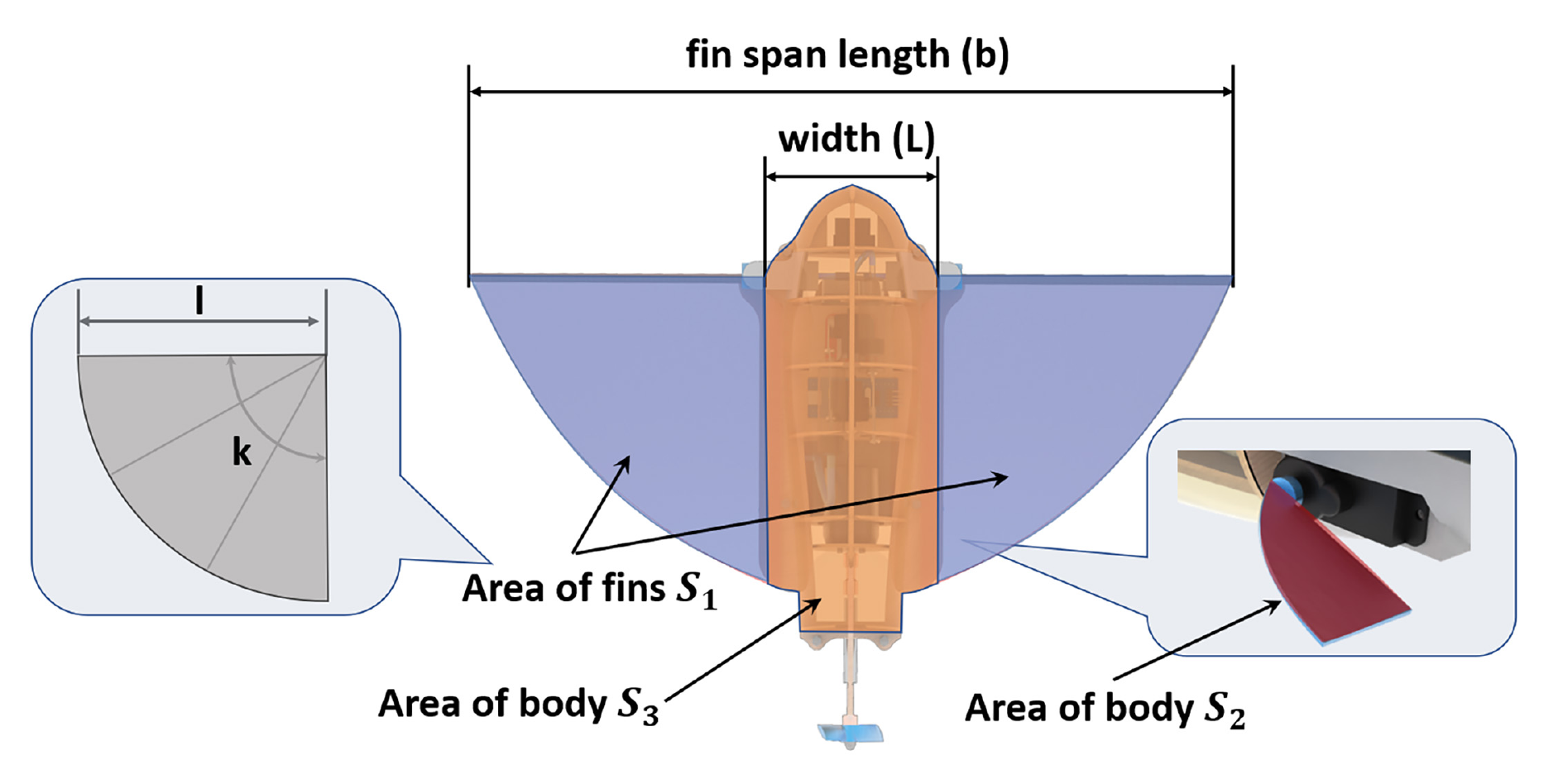}
\caption{Vertical projected area.}
\label{area}
\vspace{-5 mm} 
\end{figure}

The calculation formula of flying fish pectoral fin area $S_{1}$:
\begin{equation}
    S_{1}=\int_{0}^{k}\frac{1}{2}l^{2}dk
\end{equation}
%$K$ is the expansion angle of the pectoral fin.$l$ is the length of the pectoral fin. $c$ is the average aerodynamic chord length of flying fish, $b$ is the fin span of robot:

$k$ is the expansion angle of the pectoral fin, and the area of the pelvic fin is $S_{2}$. $l$ is the length of the pectoral fin. $c$ is the average aerodynamic chord length of flying fish, $b$ is the fin span of robot:
\begin{equation}
    b = 2l sink + L,
\end{equation}
where $L$ is is the width of the robot without wings.

In the gliding phase, the force F applied on the flying fish comes from the gravity G, aerodynamic force on the pectoral fin $F_{m}$ and pelvic fin $F_{P}$:
\begin{equation}
F=G+F_{m}+F_{p}
\end{equation}

Gravity represents in the body frame as:
\begin{equation}
    G=R_{b g}\left[\begin{array}{ccc}
      0 \\
      0 \\
      m g   
    \end{array}\right]
=m g\left[\begin{array}{ccc}
     -S_{\theta} \\
C_{\theta} S_{ \emptyset} \\
C_{\theta}  S_{\emptyset}
\end{array}\right]
\end{equation}

Both aerodynamic forces in the body frame can be showed:

\begin{equation}
F_{m}=S_{a b}{ }^{-1}\left[\begin{array}{ccc}
    -D_{m} \\
Y_{m} \\
-L_{m}
\end{array}\right]=\frac{1}{2} S_{a b}{ }^{-1} \rho V_{m}^{2} S_{1}\left[\begin{array}{ccc}
   -C_{D} \\
C_{Y} \\
-C_{L}  
\end{array} \right]
\end{equation}

\begin{equation}
F_{p}=S_{a b}{ }^{-1}\left[\begin{array}{ccc}
   -D_{p} \\
Y_{p} \\
-L_{p}  
\end{array}\right]=\frac{1}{2} S_{a b}{ }^{-1} \rho V_{p}^{2} S_{2}\left[\begin{array}{ccc}
-C_{D} \\
C_{Y} \\
-C_{L}
\end{array}\right]
\end{equation}

D, Y, and L are resistance, side force and lift, expressed in the airflow coordinate system; $C_{D}$, $C_{Y}$, $C_{L}$ are drag, side force, and lift coefficients, respectively.

\subsubsection{Conversion}
The posture and movements during water exit are key factors affecting the aerial motion of the robot. A thorough analysis and modeling of the water exit process can provide sufficient theoretical basis for the subsequent control of the robot. Before the robot exits the water, the collapsible pectoral fins are not deployed, and the direction of the body's velocity is basically the same as the direction of the body. Here, it is assumed that the velocity coordinate system and the body coordinate system remain consistent in direction at first stage during the transition phase of water exit.

\begin{figure}[h]
\centering
\includegraphics[width=8.6cm]{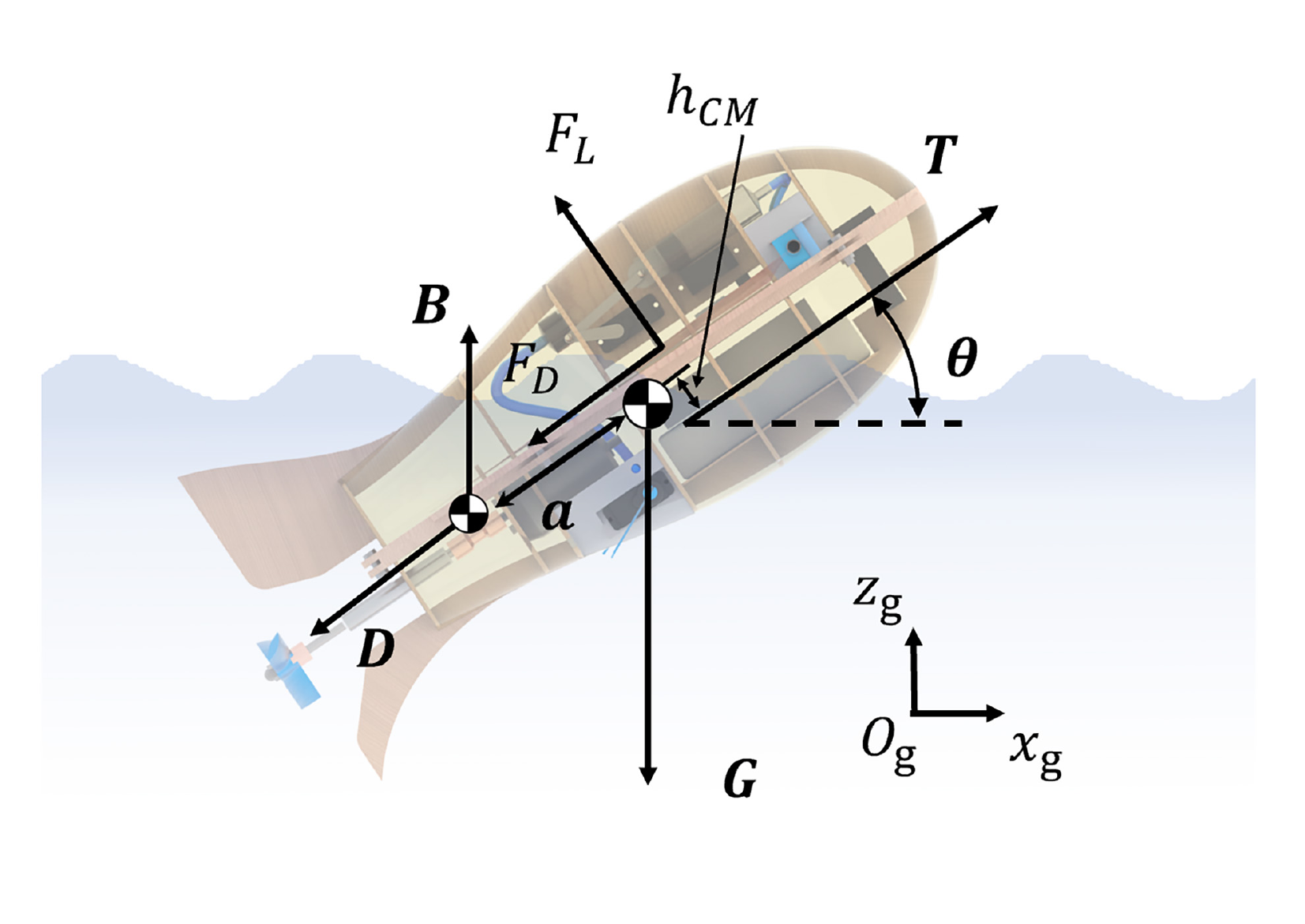}
\caption{Dynamic analysis of water exit.}
\label{exit}
\vspace{-5 mm} 
\end{figure}
During the water exit process, the robot as a whole is subject to gravity $G$ and thrust $T$ generated by the propeller rotation showen by Fif.\ref{exit}. The pectoral fins are deployed, producing aerodynamic lift $F_{L}$. The underwater part is subject to buoyancy $B$, drag forces $D_{w}$ from horizontal velocity, and viscous drag forces $F_{w}$ generated by the robot's surface movement in the water. At the same time, as the center of force of the underwater part changes with the degree of water exit, a pitching moment $M_{b}$ is generated, which affects the pitching angle state. 
\begin{equation}
    F=G+F_{L}+F_{D}+B+T+F_{w}
\end{equation}
\begin{equation}
    M=M_{m}+M_{B}+M_{T}+M_{w}
\end{equation}
The buoyancy formula and the buoyancy moment in the body coordinate system are:
\begin{equation}
B=S_{bg}\left[\begin{array}{ccc}
0\\ 
0\\ 
\rho _{w}gV_{w}
\end{array}\right], M_{B} = a\cos(\theta)\left[\begin{array}{ccc}
0\\ 
B\\ 
0
\end{array}\right]
\end{equation}
Here, the body is approximated as a cylindrical shape, and the volume of the body submerged in water is calculated based on the height of the center of gravity of the body from the water surface.
\begin{equation}
    V_{m} =\Delta z\pi R^2\sin(\theta) 
\end{equation}
where $R$ calculated as 20mm. The hydrodynamic moment and the thrust torture are
\begin{equation}
M_{w}=\frac{1}{2}\rho _{w}v_{w}^2AL_{B}\left[\begin{array}{ccc}
C_{x}\\ 
0\\ 
C_{z}
\end{array}\right], M_{T} = T \times h_{CM}
\end{equation}
where 
\begin{equation}
    F_{w}=S_{ab}^{-1}\left[\begin{array}{ccc}
-D\\ 
0\\ 
-L
\end{array}\right]
=\frac{1}{2}\rho _{w}v_{w}^2\left[\begin{array}{ccc}
C_{Dw}\\ 
0\\ 
C_{Lw}
\end{array}\right]
\end{equation}
Where 
\begin{equation}
    D=\frac{1}{2}\rho _{w}v_{w}^2AC_{Dw}
\end{equation}
\begin{equation}
    L=\frac{1}{2}\rho _{w}v_{w}^2AC_{Lw}
\end{equation}
$\rho_{w}$ is the density of the fluid, $V_{w}$ is the volume of the flying fish, $v_{w}$ is the speed of the flying fish in water, $A$ is the lateral area of the flying fish.
\begin{figure}[h]
\centering
\includegraphics[width=8.5cm]{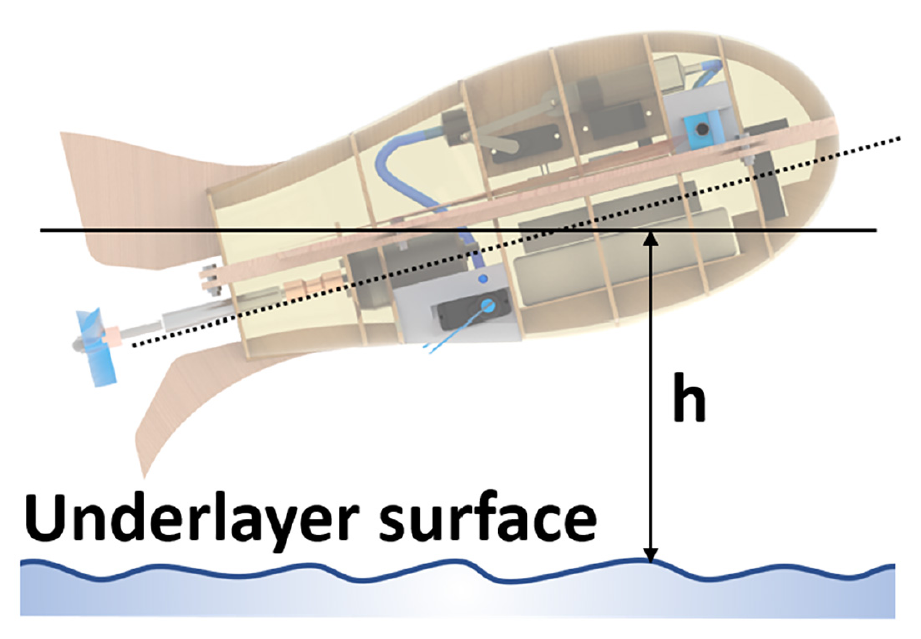}
\caption{Clearance between the robot and and underlying surface}. 
\label{grounda clearance}
\vspace{-5 mm} 
\end{figure}

\subsubsection{Ground effect}
Based on previous research on the motion pattern of flying fish, they fully utilize the ground effect to enhance lift during long-distance gliding\cite{rf18}. The ground effect refers to the phenomenon where the wing surface near a underlying surface enhances the ratio of lift and drag coefficient which means the robot can move longer distance than far away the surface. Lift coefficient during flight increases rapidly as it approaches the surface. In this process, the ratio of the distance from an object to the medium surface to the chord length, $h/b$ , is used to measure the proximity of different objects. The clearance between the robot and underlying surface is shown in Fig.\ref{grounda clearance}.  When it is less than 1/2, the impact of ground effect on lift is extremely significant, and the calculation formula for lift coefficient will also take into account the effect of ground effect\cite{rf19}. $C_{gel}$ and $C_{ged}$ represent the effect coefficients of ground effect to the lift and drag. 
\begin{equation}
    \eqalign{L=\frac{1}{2}\rho _{w}v_{w}^2AC_{Lw}C_{gel}\\
    D=\frac{1}{2}\rho _{w}v_{w}^2AC_{Dw}C_{ged}}
\end{equation}

\section{Experimental Evaluation}
To achieve precise control of the collapsible wings for the further study, several experiments were conducted to test the soft hydraulic actuator. The relationship between the bending angle and the internal pressure or volume of injected liquid was examined to establish a direct correlation. Besides, aerodynamic coefficients were also tested to ensure the locomotion of the robot can be simulated. Basing on these pretests, three experiments about discharge angle, area of collapsible wings and ground effect are finished to explore the performance of collapsible wings in the aquatic-aerial motion.

\begin{figure*}[h]
\centering
\includegraphics[width=15cm]{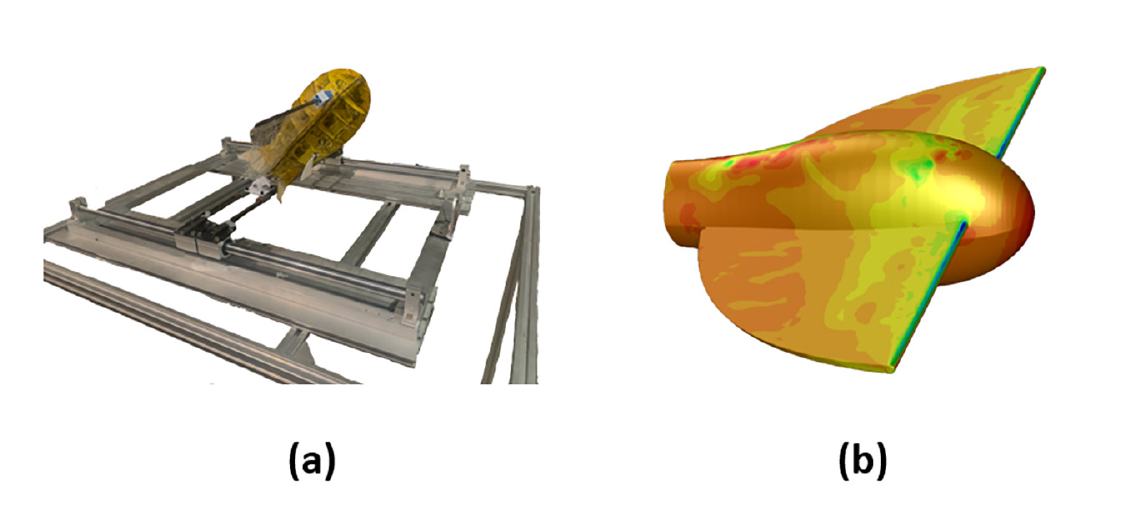}
\caption{Experiments for aerodynamic coefficients}. 
\label{experiment2}
\vspace{-5 mm} 
\end{figure*}

\subsection{Soft Actuator}
 The position of collapsible wings are controlled by the hydraulic system. As the liquid is injected or sucked out, the deformation angle of the flexible actuator changes with the internal pressure. Obtaining the correspondence between the deformation angle and the internal pressure or volume is necessary for controlling the collapsible wings. However, due to the deformation characteristics of the material itself of the flexible structure, the relationship between the two is hard to obtain by numerical calculation or simulation. Here, the angle of deformation under different pressures is recorded experimentally to measure the correspondence. Four sets of experiments were designed using liquid and air as the fillers for the flexible actuator, respectively. The internal pressure values were recorded at multiple states during the deformation from 0 to 90 degrees.

\begin{figure*}[t]
\centering
\includegraphics[width=17cm]{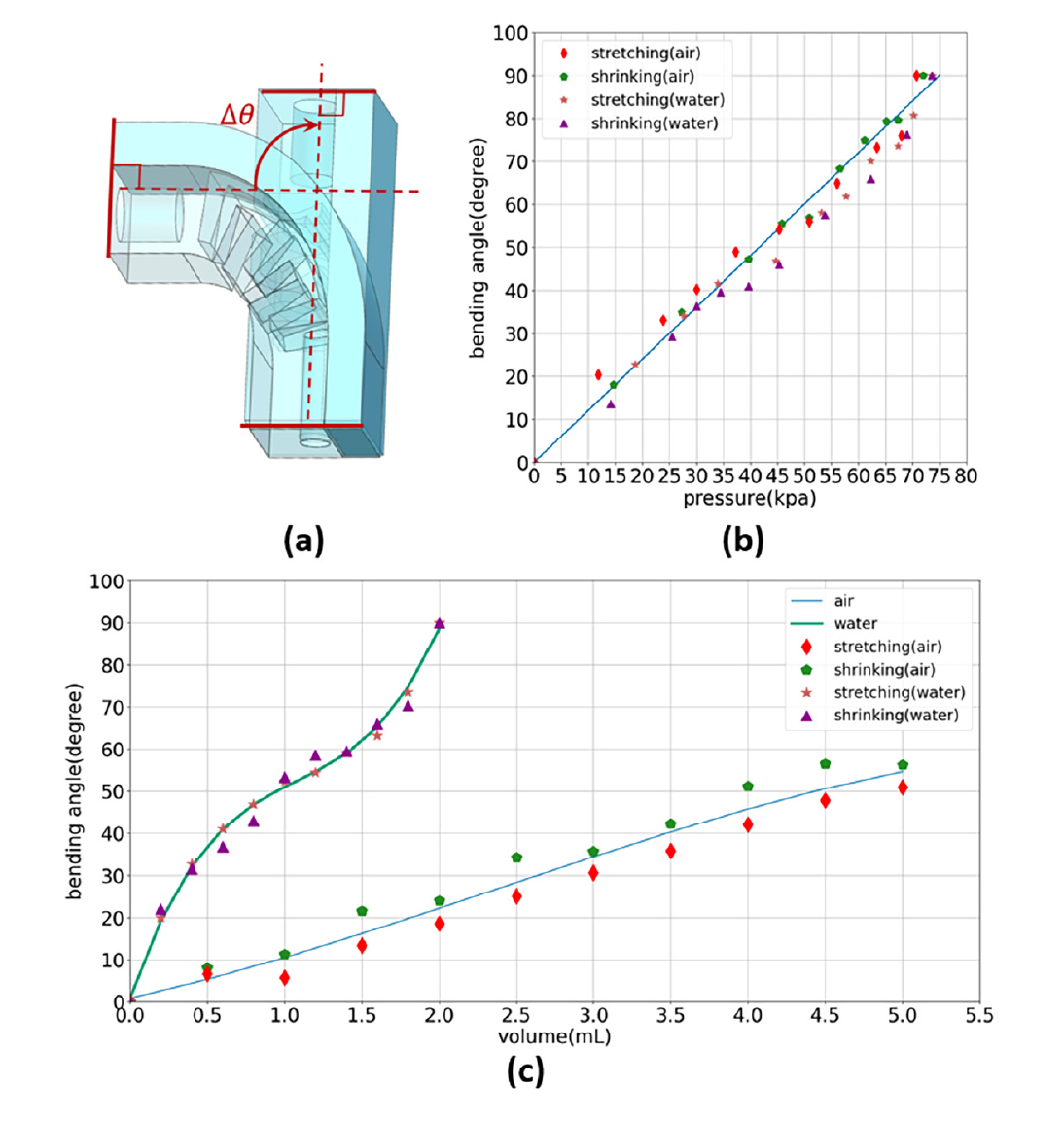}
\caption{\textbf{(a)} Bending angle of soft actuator; \textbf{(b)} Bending angle vary with the pressure; \textbf{(c)} Bending angle avry with the volume. }. 
\label{experiment1}
\vspace{-5 mm} 
\end{figure*}

Based on the data from the four sets of experiments, the relationship between the internal pressure and the deformation angle was fitted using the least-squares method, resulting in the graph shown above. It can be seen that the deformation angle varies linearly with the increase of pressure both in two medium. The accurate bending angle $\theta$ can be calculated by the equation:
 \begin{equation}
    \theta = 1.2*P,
\end{equation}
where P is the internal pressure. The four sets of experiments had good consistency without significant differences due to different medium. However, for the method of bending the actuator by injecting medium, volume is the direct control parameter. Therefore, it is more important to obtain the relationship between the injected medium volume and the deformation angle instead of the pressure. Two sets of experiments were conducted separately, and the results are shown in Fig.\ref{experiment1} (c). The bending angles of the actuator were measured for two different medium when different volumes injected during the expansion and contraction processes. Comparing gas and liquid medium, it was found that only 2 $mL$ of liquid was needed to completely drive the actuator extension, which was 40\% of volume when gas was used as the medium. The more volume means the larger hydraulic system, which is not suitable in such a micro robot. For these reasons, liquid is a better choice as the medium in this situation. In addition, the reset precision and natural affinity to water environments of the hydraulic soft actuator are more suitable for aquatic-aerial robot\cite{rf17}.

\begin{figure*}[h]
\centering
\includegraphics[width=17cm]{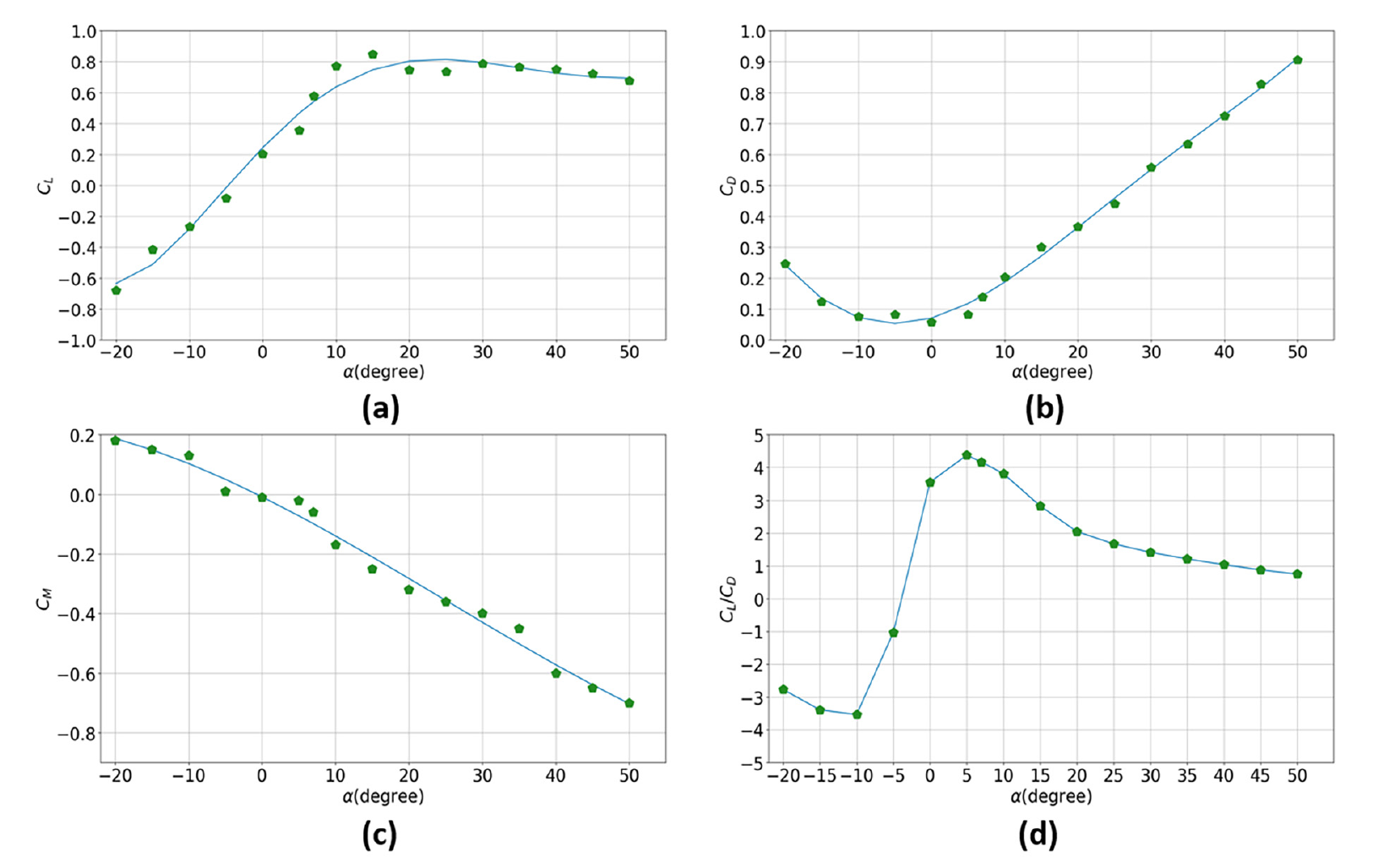}
\caption{Relationship between aerodynamic coefficients and atattck angle .\textbf{(a)} \textbf{(b)} \textbf{(c)} show variations of the lift($C_{L}$), drag($C_{D}$) and torture($C_{M}$) coefficients with the attack angle respectively. \textbf{(d)} is the ratio of lift and drag in different attack angles}. 
\label{coefficients}
\vspace{-5 mm} 
\end{figure*}

\subsection{lift coefficient}
 During stages of takeoff and gliding in the air, lift force is mainly provided by open collapsible wings. For the motion control of robots in the air, the relevant aerodynamic coefficients of the robot are crucial, including lift coefficient, drag coefficient, and torque coefficient. These coefficients are related to the robot's motion speed, the vertical projection area of the robot, and the attitude of the body. To obtain these motion coefficients for better control of the robot, an experiment was designed as shown in Fig.\ref{experiment2}. The robot was installed on a two-dimensional movable workbench, and a wind with a speed of 10m/s was applied at different angles of attack. The forces acting on the robot in the upward $F_{z}$ and backward $F_{x}$ in the ground coordinate system were measured. The lift force $L$ and drag force $D$ at different angles of attack were then calculated using these two direction forces and the angle of attack, and the calculation formula is eq.\ref{experiment2e}.

\begin{equation}
    \eqalign{F_{z}=L*C_{\alpha}+D*S_{\alpha}\\
    F_{x}=L*S_{\alpha}-D*C_{\alpha}}
    \label{experiment2e}
\end{equation}
However, since physical experiments cannot measure the rotational torque of the robot during motion under limited condition, the finite element fluid simulation method was chosen to measure the rotational torque of the robot at a speed of 10m/s showed Fig.\ref{experiment2}(b). Based on the prototype measurement, the center of gravity of the model was set at a position 13cm away from the head. After obtaining the lift, drag, and torque at each angle of attack, the robot's coefficients are calculated using the following formulas.

\begin{equation}
    \eqalign{C_{L}=\frac{2L}{\rho S V^2}\\
    C_{D}=\frac{2D}{\rho S V^2}\\
    C_{M}=\frac{2M}{\rho S V^2}}
\end{equation}

The relationship between lift coefficient and drag coefficient and angle of attack is shown in Fig.\ref{coefficients}. At small angles of attack, the lift coefficient increases linearly with the angle of attack. It tends to stabilize at 0.8 after the angle of attack exceeds 10 degrees. The drag coefficient increases with the angle of attack when it is positive and decreases with the angle of attack when it is negative. The drag coefficient is minimum at around -5 degrees angle of attack. Normally, for a typical symmetrical airfoil, the angle of attack where the lift coefficient is 0 and the angle of attack where the drag coefficient is minimum are both at 0 degrees angle of attack. The reason why they are at around -5 degrees angle of attack in this case is a 5-degree angle with respect to the horizontal plane exist in the design, inspired by the characteristic of flying fish pectoral fins. The torque coefficient decreases continuously as the angle of attack increases. A positive torque coefficient indicates that the torque of the robot's head tilts upward, while a negative torque tilts it downward. When the angle of attack is less than 0, the torque coefficient is positive, and the body tilts upward. When the angle of attack is greater than 0, the torque is negative, and the body tilts downward. Negative feedback control of the torque ensures that the body posture of the robot is consistent with its direction of motion, maintaining the stability of its movement.
The lift-to-drag ratio is an important indicator to measure the flight performance of a winged robot, and a higher lift-to-drag ratio indicates better flight performance. Fig.\ref{coefficients}\textbf{(d)} shows the relationship between the lift-to-drag ratio and the attack angle of the robot. Starting from 0, as the attack angle increases, the lift-to-drag ratio increases rapidly, reaches the maximum value 4.37 at $\alpha=5^{\circ}$, and then slowly decreases. 

\begin{figure*}[t]
\centering
\includegraphics[width=17cm]{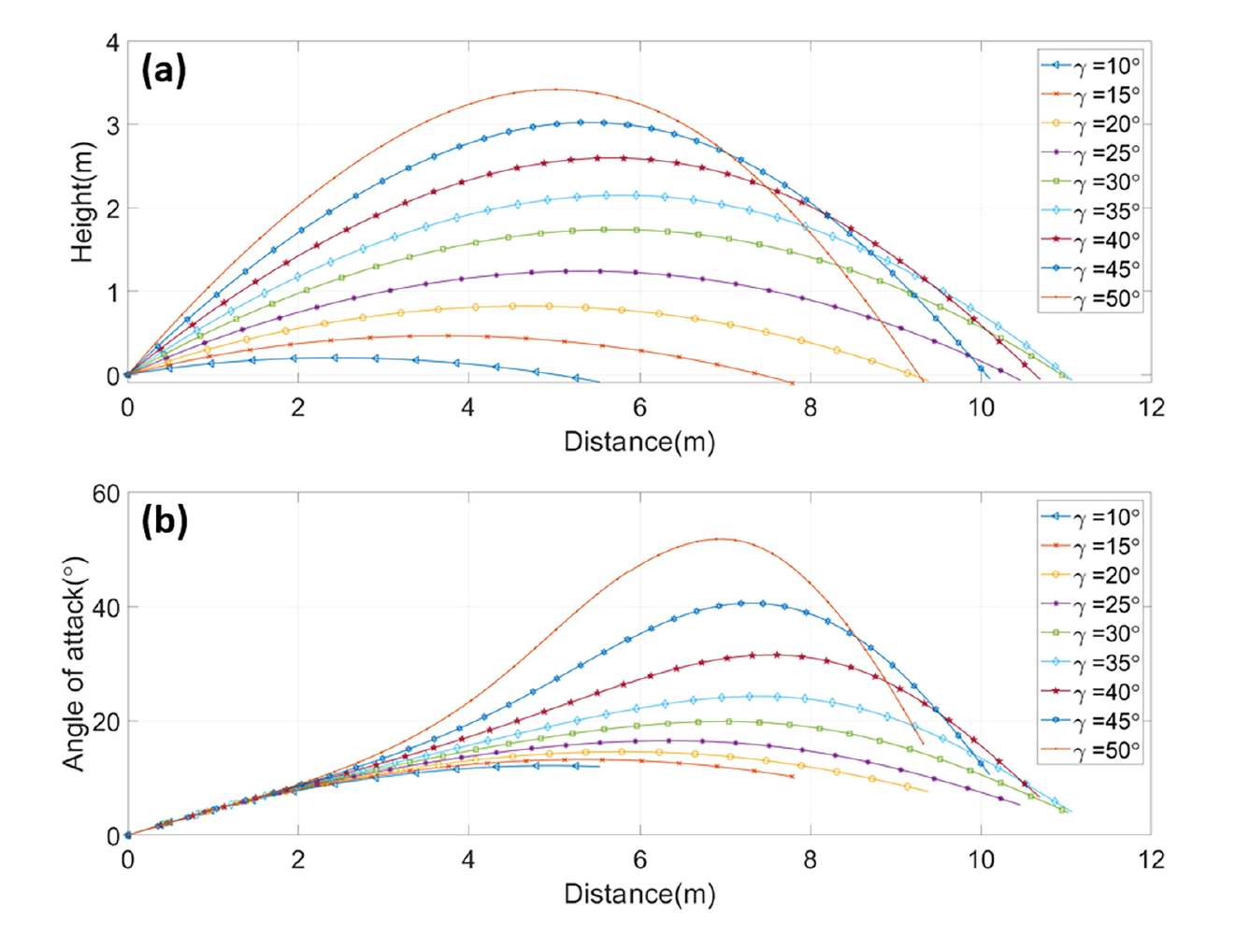}
\caption{Simulation of the motion of aquatic-aerial robot in different incline angles. \textbf{(a)} show the routine in the x-z plane; \textbf{(b)} show the variation of attack angle with the distance}. 
\label{experiment32}
\vspace{-5 mm} 
\end{figure*}

\subsection{Simulation of the locomotion with collapsible wings}
Flying fish evolve a unique motion pattern to escape predators. In the water, there are many pursuers like sword fish, tuna and other bigger fish. Although flying fish has a higher maximum speed than the pursuers, the high speed is hard to sustained for a long time in the fierce chasing. Flying fish evolve the capacity to break out of the water and hold a long time and distance with the expanding large pectoral fins over the surface, where the the predators can not touch. Even if the bigger fish try to jump out of the water, the heavy body drop it down the water rapidly. Flying fish are able to escape the predators in the water by pectoral wings in this way. However, the gliding flying fish turn to the hands-down delicious food for the hunter in the air, like seabird. For the fish, the gliding process in the air is clumsy to change the trajectory usually. Fortunately, the adjust of gesture with the pectoral wings control the flying fish motion in the air. When Flying fish are chased by the seabird, they fold their fins and drop down to the water speedy. Flying fish switching in the water and air frequently with the function of collapisble pectoral wings when chased by predators. 

Certainly, the unique motion pattern with collapsible fins is not easy even for the flying fish. Some flying fish fly too high with a wrong discharge angle lead to the balance losing and be caught by the seabirds. Appropriate discharge angle is significant and challenging especially in the roaring waves. That's the reason why the first part of simulation focus on the discharge angle. In addition, the area of opened pectoral fins is also the significant element effecting the motion of the flying fish. They reduce the area of the fins to reduce the lift and drop back to the water faster. Besides, flying fish is an intelligent creature to take full advantage of the environment. The large area of pectoral fins are used to take advantage of ground effect, which can generate a higher lift, to extend the time and distance of gliding. In this section, we also simulate this two motion to evaluate the effect of the collapsible wings.

\begin{figure*}[t]
\centering
\includegraphics[width=16cm]{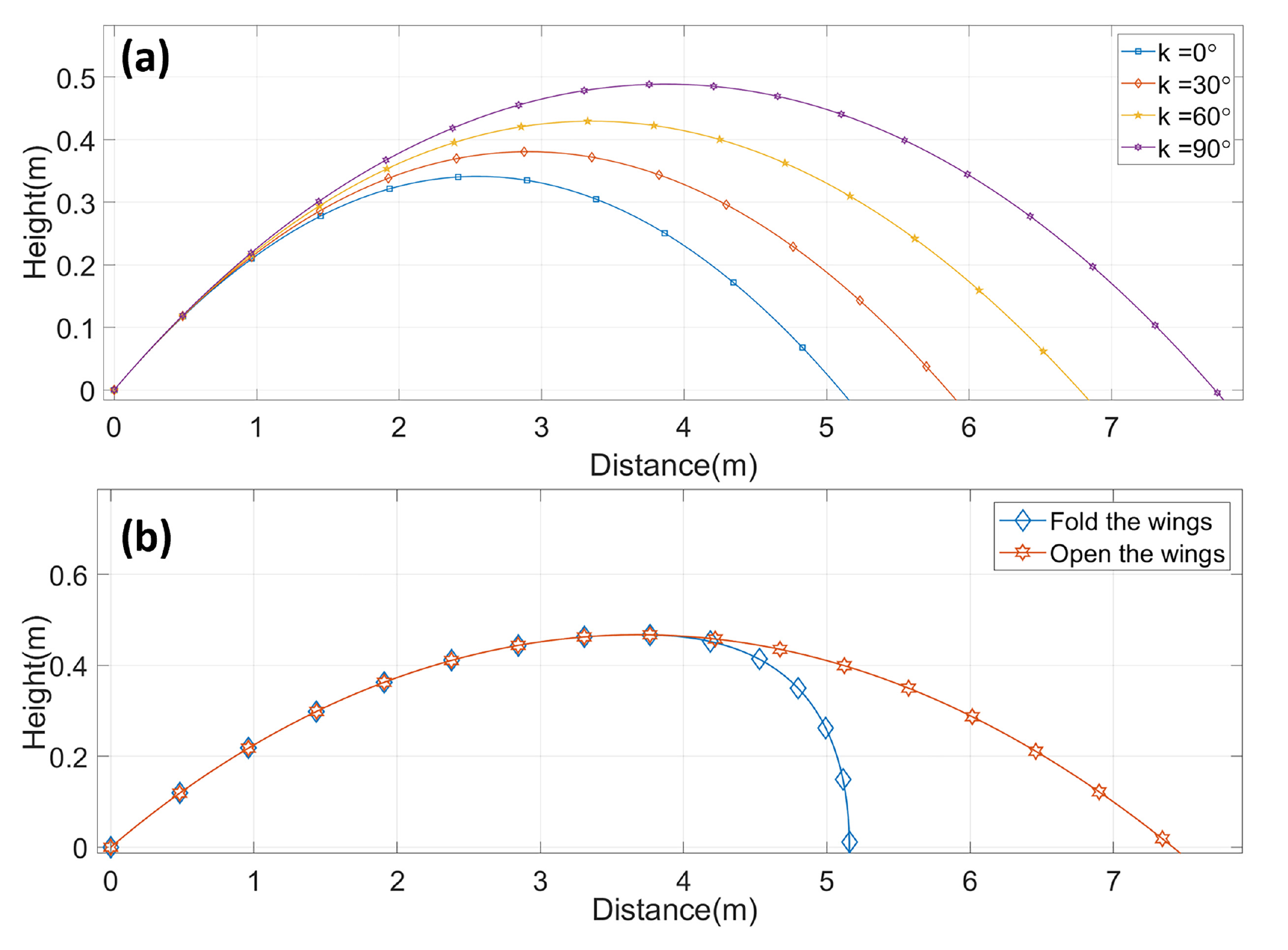}
\caption{\textbf{(a)}is the locomotion routine with different open angles wing in x-z plane when discharge angle is $15^{\circ}$ and velocity is 10$m/s$.\textbf{(b)} shows the trajectory of bionic robot with open wings and fold wings at the middle of process.}
\label{experiment33}
\vspace{-5mm} 
\end{figure*}
\subsubsection{Discharge angle effect performance of collapsible wings}
The base function of collapsible wings is generate lift force to accomplish the motion in the air. Different discharge angles lead to different trajectories with the collapsible wings. An exist research introduces the flying fish can glide for a distance up to 48m and a height up to 10m\cite{rf21} and another get the maximum gliding distance is 45.4m with a initial velocity of 20m/s from a series simulations with different discharge angles\cite{rf22}. It's hard to perform perfectly each time especially chased by predator in the water. We curious about the motion capacity of the bionic robot and simulate the moving process over the water surface in different discharge angles. Here, 11 groups of experiments with open wings conducted in intervals of 5 degrees from 10 degrees to 50 degrees with a initial velocity of 10$m/s$. The results are shown in Fig.\ref{experiment32}.

%{The influence of the angle of the water outlet on the movement of the robot after the water is very significant, and many existing amphibious robots have conducted a series of experiments on this\cite{rf20}. However, due to the different design of each robot structure, the lift coefficient, drag coefficient, and torque coefficient of the robot are different, which makes the effect of the angle of the water outlet on the post-discharge motion state of the robot inconsistent. Here, we studied the simulation of the water outlet movement state in different incline angles, with 11 groups of experiments conducted in intervals of 5 degrees from 10 degrees to 50 degrees. The results are shown in Fig.\ref{experiment32}.%}

\begin{figure*}[t]
\centering
\includegraphics[width=17cm]{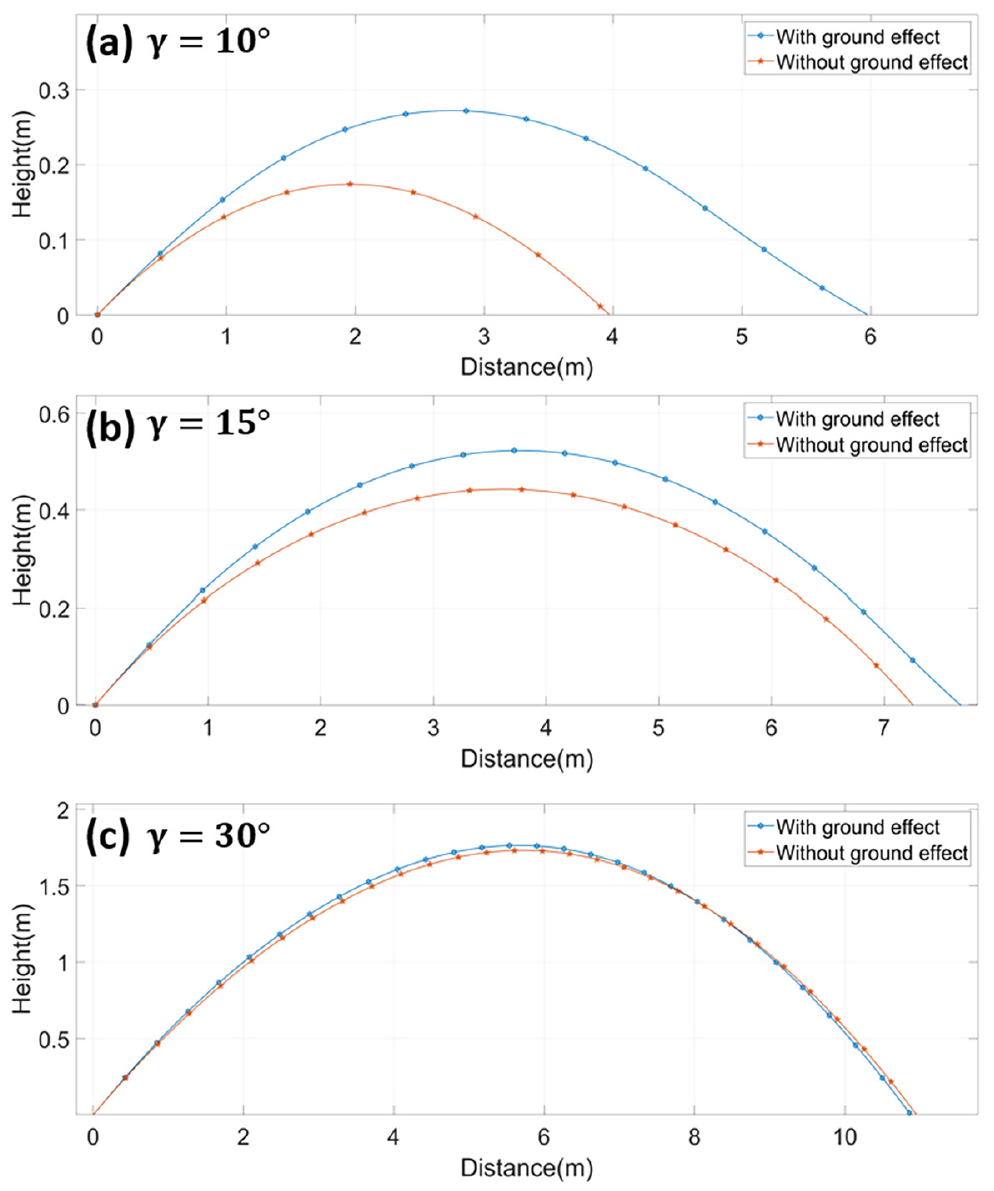}
\caption{Comparison between the motion with and without ground effect.}. 
\label{experiment34}
\vspace{-5 mm} 
\end{figure*}

In Fig.\ref{experiment32}(a), the motion trajectory of the robot with collapsible wings in the x-z plane after water discharge is shown. As the discharge angle increases, the gliding distance first increases and then decreases, which is similar to the truly presentation of the flying fish. When the angle of the water discharge is 35 degrees, the robot can glide the farthest distance, approximately 11 meters, at a flight altitude below 2.3 meters. When the water discharge angle is 50 degrees, the highest altitude is reached, approaching 3.5 meters. In figure (b), the variation of the angle of attack of the robot in the air at different distances is shown. At the initial stage, the angle of attack variation is similar for different incline angles, and it begins to vary after 3 meters. The larger the water discharge angle, the more drastic the change in the angle of attack after 3 meters, which is due to the faster change in the velocity incline angle, while the change in body attitude is relatively stable. As the angle of attack rises to its highest point, according to the determination of the torque coefficient in experiment two, a larger reverse torque causes the body to return to the direction close to the velocity incline angle, which reduces the angle of attack in the latter part of the gliding process. Not too large attack angle ensure the stability of the robot with collapsible wings, because once the attack angle is too large, the stall happens. Stall leads to loss of control easily even to the flying fish. The outcome explain the robot is able to finish the air motion part like flying fish, though the distance and height is not better than the real creature. A more lightweight and miniaturized design or a larger collapsible wings may the method to enhance the flying ability to the robot. 

\subsubsection{Motion on different angles of collapsible wings }
Not like the other fish or flying squid dropping into water immediately without flexible controlling after breaking out of the water surface, flying fish's huge pectoral fins open to glide for a long distance. The main factor contribute to the motion in the air of flying fish is the collapsible fins. We curious about the extent angle of the wings affect the movement. Thus, at an exit angle of 15 degrees and velocity at 10$m/s$, the motion of the robot under different pectoral fin opening angles was simulated. The four scenarios were 0 degrees, 30 degrees, 60 degrees, and 90 degrees of fin opening. Besides, flying fish fold their pectoral fins and adjust their gesture to drop rapidly when they chases by predator in the air. The agile action also relate to the use of the pectoral fins. A simulation about the trajectory of robot folding the collapsible wings when it touch the highest point is set and compare to the trajectory without folding action.   

Fig.\ref{experiment33} shows the experimental results.  (a) indicates that with a larger area of the pectoral fins, the flying distance was significantly increased. Without opening the pectoral fins, the robot could only glide 5.3 meters based on inertial motion. However, when the pectoral fins were fully open at 90 degrees, the robot glided 7.8 meters, representing a 47.2\% increase in gliding distance. And this was achieved without any control of the robot's flight attitude. The fly process with open wings not in a control already induce a significant improvement, which verify the feasibility to fly far away with further precise control. (b) shows the perform difference between the open and fold wings two actions after the highest point. After folding the wings, robot drop into the water rapidly with a moving distance about 1m. Comparing to the distance of 3.5m with a perpetual opening wings, nearly two third motion path are avoided. The less distance means less chance for predators to catch up. These two simulations verified the further precise control of the collapsible wings is meaningful and can definitely improve the locomotion capacity of the robot.

\subsubsection{The effect of Ground effect}
Flying fish is smart to take advantage to the environment especially in the use of ground effect for a longer gliding distance. The ground effect is the reason why flying fish evolved to glide at a close distance to the water surface for long distances with extending pectoral fins. It provides a lift coefficient by collapsible wings which is more than in normal flight conditions, and this amplification increases rapidly as the clearance from the ground decreases. To verify the ground effect for the collapsible wings and lay the groundwork for future control, simulations were conducted for two scenarios: one with ground effect and one without, under the condition of different water discharge angles and 90-degree collapsible wings expansion. The initial velocity of all experiments is 10$m/s$.  Fig.\ref{experiment34} shows the results of the experiment. (a), (b), and (c) show the simulation experiments under three scenarios with water discharge angles of 10 degrees, 15 degrees, and 30 degrees, respectively. When the water exit angle is 10 degrees, the robot with ground effect can glide for 6 meters, while the one without can only glide for 4 meters, resulting in a 50\% increase in distance. When the water exit angle is 15 degrees, the ground effect gain decreases to 0.5 meters, resulting in a 6.8\% increase, and when the water exit angle is 30 degrees, the ground effect does not provide an increase in gliding distance, even decreases it, mainly due to the negative impact of ground effect when exiting the water. At the same time, the flying height is also of concern. Since the ground effect has a significant effect only when the clearance $h$ between the collapsible wings and ground is less than half of the wingspan. For the robot with a wingspan of 0.5, the ground effect only has an evident effect when the clearance above the ground is below 0.25 meters. For a low water exit angle of 10 degrees, the gliding height is just within this range, so the ground effect provides a huge boost to the gliding distance. As the flying height increases, the proportion of distance traveled below 0.25 meters gradually decreases throughout the process, and the effect of ground effect to collapsible wings becomes smaller. The experiments proved the method of collapsible wings to use the ground effect for a longer moving distance.

\subsection{Discussion}
With the coefficients tested, experiments simulate the moving process of robots with collapsible wings in the difference discharge angle, explore the effect by area of collapsible wings and the collapsible wings in ground effect. In addition, these experiments verify the motion pattern of taking off, flying and gliding is available for the robot with collapsible wings. As the outcome of the simulation, bionic robot is able to control the area of the collapsible wings to adjust the gesture and statement of moving process. Furthermore, the use of ground effect definitely increase the flying capacity near the water surface. Taking advantage of the ground effect is an excellent way to extend the endurance in the future. This motion pattern truly a outstanding strategy with collapsible wings for the aerial aquatic robot.  

\section{Conclusions}
Inspired by the unique motion pattern with collapsible wings, this paper explore the advantage of the collapsible wings in the aquatic-aerial robots. A bionic robot prototype with collapsible wings is designed and produced to test the coefficients of locomotion. After the dynamic modal of the robot is built, several experiments has been test to verify the advantage of collapsible in the multi-modal motion pattern. 

This article's main contribution includes designing aquatic-aerial robot with collapsible wings, which use the soft hydraulic actuator, and introduce the design and fabrication of the soft actuator structure. Dynamic model of the robot is built and the study experimentally and computationally analyzed various dynamic parameters of the robot, including lift coefficient, drag coefficient, and torque coefficient at different attack angles. Finally, all the information pretested were used to simulate and analyze the motion state of the flying fish bio-robot with collapsible wings. The experiments concentrate to the discharge angle, area of the collapsible wings and the use of ground effect. Collapsible wings in the aquatic-aerial robots truly contribute to the unique motion pattern of the flying fish and the future aquatic-aerial robots. These outcomes of collapsible wings laying a necessary theoretical foundation for future work.

\section*{Acknowledgments}
This work was supported by the National High Technology Research and Development Program of China under Grant No. 2018YFE0204300, and the National Natural Science Foundation of China under Grant No. U1964203, and sponsored by Meituan and Tsinghua University-Didi Joint Research Center for Future Mobility.

\section*{Reference}

\bibliographystyle{iopart-num.bst}
\bibliography{iopart-num}

\providecommand{\newblock}{}
\begin{thebibliography}{10}
\expandafter\ifx\csname url\endcsname\relax
  \def\url#1{{\tt #1}}\fi
\expandafter\ifx\csname urlprefix\endcsname\relax\def\urlprefix{URL }\fi
\providecommand{\eprint}[2][]{\url{#2}}

\bibitem{rf1}
Siddall R and Kovac M 2016 {\em IEEE/ASME Transactions on Mechatronics\/} {\bf
  22} 217--226

\bibitem{rf2}
Zimmerman S and Abdelkefi A 2017 {\em Progress in Aerospace Sciences\/} {\bf
  93} 95--119

\bibitem{rf3}
Zeng Z, Lyu C, Bi Y, Jin Y, Lu D and Lian L 2022 {\em Ocean Engineering\/} {\bf
  248} 110840

\bibitem{rf4}
Weisler W, Stewart W, Anderson M~B, Peters K~J, Gopalarathnam A and Bryant M
  2017 {\em IEEE Journal of Oceanic Engineering\/} {\bf 43} 969--982

\bibitem{rf5}
Rockenbauer F~M, Jeger S, Beltran L, Berger M, Harms M, Kaufmann N, Rauch M,
  Reinders M, Lawrance N~R, Stastny T {\em et~al.\/} 2021 Dipper: A dynamically
  transitioning aerial-aquatic unmanned vehicle. {\em Robotics: Science and
  Systems\/} pp 12--16

\bibitem{rf6}
Li L, Wang S, Zhang Y, Song S, Wang C, Tan S, Zhao W, Wang G, Sun W, Yang F
  {\em et~al.\/} 2022 {\em Science Robotics\/} {\bf 7} eabm6695

\bibitem{rf7}
Hu R, Lu D, Xiong C, Lyu C, Zhou H, Jin Y, Wei T, Yu C, Zeng Z and Lian L 2022
  {\em Applied Ocean Research\/} {\bf 120} 102925

\bibitem{rf8}
Yang X, Wang T, Liang J, Yao G, Chen Y and Shen Q 2012 Numerical analysis of
  biomimetic gannet impacting with water during plunge-diving {\em 2012 IEEE
  International Conference on Robotics and Biomimetics (ROBIO)\/} (IEEE) pp
  569--574

\bibitem{rf12}
Siddall R, Ortega~Ancel A and Kova{\v{c}} M 2017 {\em Interface focus\/} {\bf
  7} 20160085

\bibitem{rf13}
Siddall R and Kova{\v{c}} M 2015 A water jet thruster for an aquatic micro air
  vehicle {\em 2015 IEEE International Conference on Robotics and Automation
  (ICRA)\/} (IEEE) pp 3979--3985

\bibitem{rf14}
Hou T, Yang X, Su H, Jiang B, Chen L, Wang T and Liang J 2019 Design and
  experiments of a squid-like aquatic-aerial vehicle with soft morphing fins
  and arms {\em 2019 International Conference on Robotics and Automation
  (ICRA)\/} (IEEE) pp 4681--4687

\bibitem{rf15}
Meadows G, Atkins E, Washabaugh P, Meadows L, Bernal L, Gilchrist B, Smith D,
  VanSumeren H, Macy D, Eubank R {\em et~al.\/} 2009 The flying fish persistent
  ocean surveillance platform {\em AIAA Infotech@ Aerospace Conference and AIAA
  Unmanned... Unlimited Conference\/} p 1902

\bibitem{rf16}
Amy G and Techet A~H 2011 Design considerations for a robotic flying fish {\em
  OCEANS'11 MTS/IEEE KONA (pp. 1-8). IEEE.\/} pp 1--8

\bibitem{rf10}
Park H and Choi H 2010 {\em Journal of Experimental Biology\/} {\bf 213}
  3269--3279

\bibitem{rf18}
J D 2003 {\em Journal of Fish Biology.\/}  455--463

\bibitem{rf19}
Boschetti~PJ C~E and González~PJ M~A 2020 {\em Journal of Aircraft.\/}
  1234--1241

\bibitem{rf17}
Xavier M~S, Tawk C~D, Zolfagharian A, Pinskier J, Howard D, Young T, Lai J,
  Harrison S~M, Yong Y~K, Bodaghi M {\em et~al.\/} 2022 {\em IEEE Access\/}

\bibitem{rf21}
Davenport J 1994 {\em Reviews in Fish Biology and Fisheries\/} {\bf 4} 184--214

\bibitem{rf22}
Deng J, Zhang L, Liu Z and Mao X 2019 {\em Bioinspiration \& biomimetics\/}
  {\bf 14} 046009

\end{thebibliography}
\end{document}